\documentclass[11pt]{article}
\usepackage{graphicx}
\begin{document}
\nopagebreak[4]
\title{Optical Character Recognition, Using K-Nearest Neighbors}
\author{Wei Wang}
\maketitle
\begin{abstract}
The problem of optical character recognition, OCR, has been widely discussed in the literature. Having a hand-written text, the program aims at recognizing the text. Even though there are several approaches to this issue, it is still an open problem. In this paper we would like to propose an approach that uses K-nearest neighbors algorithm, and has the accuracy of more than 90\%. The training and run time is also very short. 
\end{abstract}
\section{Introduction}
Visual pattern recognition has been a point of interest of machine learning experts for several years. A common approach of recognizing a pattern consists of two important tasks, feature extraction and classification. As a preprocessing task, feature extraction must be done in order to find a pattern for the image. Obviously, there is no way to choose best features for a specific problem. It is usually done by some heuristics that means some different features, based on experience and problem especific knowledge, are tested on the same problem and the most efficient one is chosen. \\
As a start point for solving the problem, a list of reference objects is required in order to match any new intance to reference images. The result of the matching task is used to recognize the new object. Although finding features of an object is very helpful in recognition process, it usualy requires lots of computations and has its own complexity. Thus it is a challenging task to choose features useful and at the same time easy to be computed. \\
One of the most important reasons of choosing features instead of using pixels of an image themselves, is curse of dimensionality \cite{curse}. It is  considerable that the number of features must not be so high that increases the number of dimnesions. \\
Optical Character Recognition (OCR) is a field of research in artificial intelligence, computer vision and patern recognition that is the recognition of characters from optical data. The characters may be machine printed or handwritten \cite{ocr}. One of the aspect of OCR concerns digit recognition. Digit recognition is simpler than OCR, because of less number of possibilities and simpler format of combination of symbols. In this paper what we explain and try to solve is handwritten digit recognittion. \\
Though the problem of digit recognition has been intensively investigated, the improvement of its performance is still a major issue in a number of industrial applications, e.g. parcel sorting \cite{simple}. We here describe our approaches that uses simple methods to extract features and classify diferent digits. We developed two simillar algorithms. The first approach is a simple feature extraction and classification approach that works based on partitioning the image and averaging the partitions. The second one is very similar to the former one but gradient based.\\
The remainder of this report is organized as follows. In section two we explain two approaches that we developed and tested. In section III we show the reaults of both methods according to accuracy and timing of free run tests. The last section is the conclusion of our work. 

\section{Our Approaches}

\begin{figure}[htp]
\centering
\includegraphics[scale=0.4]{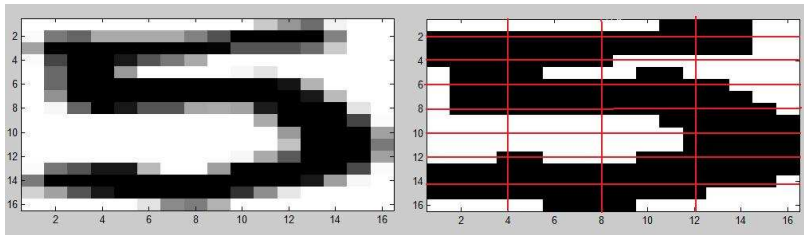}
\caption{Left: raw image, Right: Skeletonized and partitioned image}
\label{fig1}
\end{figure}

Often vision systems tend to become quite complex. We here propose a simple two-stage, and another simple one-stage method of feature extraction. We first start with two-stage method.\\
The first step of feature extraction is  skeletonization. Morphological skeleton is a skeleton representation of a shape or binary image, computed by morphological operators \cite{wiki}. Applying skeletonization, we get rid of some shadows that might disturb our approach for recognition. This approach is done by Matlab using \textit{bwmorph} recursively until there is no difference between the last two produced images. In figure \ref{fig1} the skeletonization of a digit is shown.\\
The second step of our approach is longer than the first introductory one. Receiving a thin image from the first step, we try to extract features from it. The first task is to extract the effective part of image from its whole. The image might have some useless area. This area does not store any information. This is usually a totally blank part of the image. So it is better to remove these parts as soon as possible. We name the area of meaningful data as the effective part of the image. Extracting this area, we compute the following equations to get minimum and maximum of X and Y, positions of pixels. Then based on these values we make a rectangle of efficient area, and extract features only for this area.\\
\begin{eqnarray*}
\label{eq1}
min_x = min(X,:) \mbox{ where value(X,Y) $>$ 0}\\
min_y = min(:,Y) \mbox{ where value(X,Y) $>$ 0}\\
max_x = max(X,:) \mbox{ where value(X,Y) $>$ 0}\\
max_y = max(:,Y) \mbox{ where value(X,Y) $>$ 0}
\end{eqnarray*}

Extracting features is the most important part of digit recognition. There are two considerable points in feature extraction. The first one is to avoid many number of features. Otherwise solving curse of dimensionality is neccessary. The second one is computational complexity. If feature extraction needs lots of computational power, it takes a long time to compute features for all instances. That is a very good reason for choosing simpler mathematical formulas for feature extraction. We tried two of the simplest forms of feature extraction to reach to a good timing and accuracy.\\

In the first approach we split the effective image to several rectangular sections and compute the average value of pixels for each rectangle as the representative of that section. Amongst different partitioning types, we found 32 sections more efficient than others. In this case we partitioned an effective image into 4 vertical and 8 horizontal rectangles. The results of different methods of partitioning is shown in the next section. \\
In the second approach, we do not do the skeletonization at all. Because gradient is done to find the highest intensity of black amongst different pixels of a section. Applying skeletonization makes the gradient useless.  We partitioned the image exactly the same as the previous one. Instead of computing the average value, we measured the gradient of the pixels in each section. Applying gradient to a section results in 2D list of derivatives. We choose the maximum absolute value of derivatives with respect to X as well as Y. Thus, having the same number of sections, number of features is twice of the previous algorithm. \\  

When feature extraction is done, we put the array of vectors of features in a K-Dimentional tree \cite{kd}. The kd-tree that we used was an implemented package, and we utilized it for our purpose. During the classification task, we extract first three neighbors of the input image using the kd-tree. Given a kd-tree the algorithm computes a k-nearest neighbor query (kNN) \cite{shams1, shams2} with a preprocessing time of $O(d \times N \log{N})$ and an expected query time of $O(\log{N})$ where N is number of points and d dimensionality of a point in the set.

\section{Experiment and Results}

We tested our algorithms of digit recognition on a set of 2000 instances of digit images that can be found \cite{site}. Each image was 2D, gray-scale by the size of 256X256 pixels. We trained our datamodel on 1500 samples of this dataset and tested it on the rest. It is notable that the test cases contain 50 intances of each class. The computer that we used, had a Core 2 Duo processor 1.66 GHz associated by 2 GBytes of RAM. We tested all of our cases on the same computer and showed the results in table \ref{table1}. Running software was Matlab 2008a with image processing package enabled. Image processing toolbox is required for skeletonization purpose. The Kd-tree used in this algorithm, is developed under C++ languague that needs mex and C++ compiler. We installed Microsft Visual C 2008 express edition for the compiling reason.\\

\begin{table}[t]
\centering
\begin{tabular}{c|c|c}
Name & Accuracy(percentage) & Run Time(seconds)\\ \hline
4 vertical, 4 horizontal & 87.8 & 0.795 \\ \hline
8 vertical, 4 horizontal & 86 & 1.341\\ \hline
4 vertical, 8 horizontal & 92.6 & 1.092\\ \hline
8 vertical, 8 horizontal & 91.8 & 2.152\\ \hline
Gradient Based & & \\ 4 vertical, 8 horizontal & 80.6 & 4.72 \\
\end{tabular}
\caption{Performances of different digit recognition methods}
\label{table1}
\end{table}

From the results it is easily understandable that the best performance was obtained by using 32 features when we partitioned it like figure \ref{fig1}.\\

\section{Conclusions}
In this report we explained our simple approaches of digit detection based on partitioning an image into rectangles, and using either the average pixels of sections or their maximum absolute gradients as representatives of the image, in the other words features of the image. We used kd-tree to find the nearest neighbors of an image. We showed that despite its simplicity, averaging pixel values is very efficient, esp in timing issues. \\

\end{document}